\documentclass[runningheads]{llncs}

\newif\ifanonymous
\usepackage{amsmath}
\usepackage{pifont}% http://ctan.org/pkg/pifont
\newcommand{\cmark}{\ding{51}}%
\usepackage{amssymb}
\usepackage{graphicx}

\usepackage[T1]{fontenc}
\usepackage{graphicx}
\ifanonymous
    \newcommand{\repourl}{https://github.com/anonymous}
\else
    \newcommand{\repourl}{https://github.com/Visual-Computing/MCIP}
\fi

\begin{document}
\title{Optimizing CLIP Models for Image Retrieval with Maintained Joint-Embedding Alignment}
\titlerunning{Optimizing CLIP Models for Image Retrieval}
\ifanonymous
    \author{anonymous}
\else

\author{Konstantin Schall\orcidID{0000-0003-3548-0537} \and
Kai Uwe Barthel\orcidID{0000-0001-6309-572X}
\and
Nico Hezel\orcidID{0000-0002-3957-4672}
\and
Klaus Jung\orcidID{0000-0002-3600-6848}}
\authorrunning{K. Schall et al.}
\institute{Visual Computing Group, HTW Berlin, 12459 Berlin, Germany 
\email{\{konstantin.schall,barthel,hezel,klaus.jung\}@htw-berlin}\\
\url{https://visual-computing.com/}}
\fi
\maketitle              % typeset the header of the contribution
\begin{abstract}
Contrastive Language and Image Pairing (CLIP), a transformative method in multimedia retrieval, typically trains two neural networks concurrently to generate joint embeddings for text and image pairs. However, when applied directly, these models often struggle to differentiate between visually distinct images that have similar captions, resulting in suboptimal performance for image-based similarity searches. 
This paper addresses the challenge of optimizing CLIP models for various image-based similarity search scenarios, while maintaining their effectiveness in text-based search tasks such as text-to-image retrieval and zero-shot classification. 
We propose and evaluate two novel methods aimed at refining the retrieval capabilities of CLIP without compromising the alignment between text and image embeddings.
The first method involves a sequential fine-tuning process: initially optimizing the image encoder for more precise image retrieval and subsequently realigning the text encoder to these optimized image embeddings. The second approach integrates pseudo-captions during the retrieval-optimization phase to foster direct alignment within the embedding space. Through comprehensive experiments, we demonstrate that these methods enhance CLIP's performance on various benchmarks, including image retrieval, k-NN classification, and zero-shot text-based classification, while maintaining robustness in text-to-image retrieval. 
Our optimized models permit maintaining a single embedding per image, significantly simplifying the infrastructure needed for large-scale multi-modal similarity search systems.

\end{abstract}
\section{Introduction}

Contrastive Language and Image Pairing (CLIP) \cite{CLIP} has emerged as one of the most influential developments in deep learning and similarity search in recent years. It introduces a robust joint-embedding model that excels in text-to-image similarity searches and has paved the way for pioneering work in image generation \cite{stable-difussion,DALL-E}, zero-shot object detection \cite{OWOD,detic}, and various large-scale vision foundation models \cite{ALIGN,UniCL,GLIPv2}. Additionally, CLIP has significantly advanced the field of interactive multimedia retrieval \cite{VBS12,LSC24}. The prevalence of images captioned with text on the web and social media platforms has made it possible to collect annotated images more cost-effectively and efficiently compared to traditional single-label datasets like ImageNet \cite{ImageNet}. This approach has enabled the creation of datasets on an unprecedented scale, featuring up to 8 billion image-text pairs \cite{SigLIP}. CLIP trains two neural networks concurrently: an image encoder and a text encoder, aiming to produce highly similar embeddings for corresponding image-text pairs. 
This training approach not only results in neural networks with more robust representations and improved generalization but also supports the development of architectures with a larger number of parameters \cite{ScalingViT}.

However, since CLIP employs natural text supervision, the image encoders trained by CLIP tend to perform semantic similarity searches, which are suboptimal for instance-level content-based image retrieval scenarios. 
For example, one image might show a burger and chips while another displays a fast-food restaurant building; both could be captioned with "visiting a fast-food restaurant". Consequently, an image-based search using one of the images could incorrectly return the other, despite their visual differences. A method to mitigate this issue involves specifically fine-tuning CLIP-trained image encoders for general-purpose content-based image retrieval, as described in \cite{GPR-NNS}. This fine-tuning allows the encoders to perform image-based similarity searches across diverse instance-based datasets from various domains, achieving high generalization without requiring domain-specific adaptations. However, this fine-tuning of only the image encoder can lead to a misalignment between the textual and visual embeddings, making the retrieval-optimized image embeddings less effective for text-to-image search scenarios. Applications that require good performance in both image-to-image and text-to-image retrieval \cite{VBS12,LSC24} would therefore have to maintain two separate vector sets.

This paper explores the development of joint-embedding models that perform well in both search modalities and presents two different methods to achieve a retrieval-optimized joint-embedding space. The first method involves a two-stage fine-tuning approach. Initially, the image encoder is optimized for retrieval, akin to the methods described in \cite{GPR-NNS}. Subsequently, the text encoder is realigned with the newly optimized image embeddings. The second method utilizes pseudo-captions of images to directly introduce the alignment process to the retrieval-optimization fine-tuning.  We conduct extensive experiments to highlight the strengths and weaknesses of these approaches. Moreover, we evaluate our optimized CLIP models across various benchmarks, demonstrating that our methods not only improve the original model's performance on tasks such as image retrieval, k-NN classification, and text-based zero-shot classification but also maintain high quality in text-to-image retrieval and allow for these advantages with one embedding per image. Code and optimized model weights are accessible at: \repourl.

\section{Related Work}
\subsection{Joint-Embeddings for Text and Images}

In the pre-deep learning era, Canonical Correlation Analysis (CCA) \cite{CCA} was a primary technique for joint-embedding models in text-to-image retrieval, focusing on linear correlations to align text and image vectors in a shared space. This approach, however, was limited by its linear nature and inability to grasp complex relationships.

With the advent of deep learning, the DeViSE (Deep Visual-Semantic Embedding) \cite{devise} model emerged as a significant advancement. DeViSE uses a deep neural network to project visual data from a pre-trained image model into the space of textual embeddings obtained from a language model. 

Another important deep learning model is VSE++ (Visual Semantic Embedding) \cite{VSE++}, which improved upon previous embeddings by focusing on hard negatives during training. This model is designed to enhance the discriminative power of the embeddings, making it more effective at differentiating between closely related images and texts.

CLIP \cite{CLIP} revolutionized the field by training with a contrastive objective on a large-scale collection of text-image pairs from the Internet. CLIP's ability to perform zero-shot classification, where the model can accurately categorize images it has never seen during training, demonstrates its robustness and generalization capabilities. 

Despite its strengths, CLIP's reliance on natural language descriptions can inadvertently emphasize semantic over visual similarities, often leading to discrepancies in retrieval outcomes due to the semantic-visual gap and the contextual ambiguity inherent in language when using the image embeddings for content-based image retrieval.

\subsection{Retrieval Optimizations}
Deep neural networks have significantly outperformed traditional methods like SIFT \cite{SIFT} in nearest neighbor search, offering substantial improvements \cite{NeuralCodes}. Image embedding models now generate comprehensive representations, encapsulating information across the entire image within a single global vector. For retrieval tasks, transitioning from global average pooling to advanced techniques such as RMAC \cite{RMAC} or GeM \cite{GEM} has been advantageous. Recent advancements have further refined these global image representations by incorporating local details from intermediate network layers \cite{DELG,DOLG,SuperGlobal}. Typically, these models are trained on the Google Landmarks v2 dataset \cite{GLIPv2} and evaluated against benchmarks like RParis and ROxford \cite{RParisOxford} or the Google Landmarks v2 test set. However, the effectiveness of these techniques in more generic retrieval contexts remains uncertain, as they have been primarily developed and tested within highly specific domains, where data homogeneity does not adequately represent the diversity of images encountered in general-purpose applications.

Instead of focusing on architectural modifications to improve fine-grained image detail aggregation, GPR \cite{GPR-NNS} and UnED \cite{uned} take a different approach to increase image retrieval performance in applications with a wide range of image domains. Both use a large combination of publicly available datasets with a total of over 100,000 classes to re-train CLIP image encoders for the specific case of image encoders with appropriate loss functions like the ArcMargin loss \cite{ArcFace}. This method greatly increases the desired retrieval performance, however the fine-tuned image encoders can not be used for text-to-image similarity search anymore, since the joint-embedding space became misaligned through the fine-tuning process.

\subsection{CLIP Optimizations}

CLIP models have successfully been optimized in various aspects. 
The LAION organisation published an openly available dataset of 5 billion  text-image pairs and trained a variety of large models to further improve the accuracy and performance of the original CLIP \cite{Laion5B}. 
EVA \cite{EVA} applies an iterative training procedure, heavily utilizing masked-image-modelling \cite{MAE} to be able to train very large image and text encoders with billions of parameters. SigLIP \cite{SigLIP} further improves CLIP's qualities by utilizing a novel Sigmoid based loss function for language and image pairing and trains with the \textit{webli} dataset \cite{SigLIP}, consisting of over 8 billion text-image pairs. 

Locked-Image-Tuning (LiT) \cite{lit} uses an image encoder pre-trained on a classification problem instead of a randomly initialised one and locks its weights during the CLIP training. This allows significantly faster training, since only the text-encoder has to be trained and yields better zero-shot classification performance, when the image-encoder was pre-trained with a large-scale proprietary dataset. 

RA-CLIP \cite{RA-CLIP} proposes a superior training procedure by introducing a retrieval augmented image encoder. Prior to CLIP training, image embeddings are extracted with a pre-trained network like DinoV2 \cite{DINOv2} to enhance the image embeddings of the trained image encoder with a retrieval augmentation module (RAM). 

ViCHA \cite{vicha} tries to make the training of CLIP-like models more efficient and less reliant on large-scale datasets by adding a hierarchical image-text contrastive alignment loss that compares representations across several layers of both encoders, achieving good results with only 1.1 million text-image pairs. 

However, none of these approaches try to optimize a CLIP models image encoder for effective image similarity search, while preserving the quality of text-based tasks like text-to-image retrieval and zero-shot classification.

\section{Method}

The goal of our proposed methods is to optimize CLIP-like joint-embedding models in a way that allows improved image-based similarity search with the image embeddings, while also enabling a high quality for text-to-image retrieval and text-based zero-shot classification of images.  We investigate two different approaches to achieve this goal, which will be explained in detail in this chapter after some necessary background information has been provided.

\subsection{Background}

\textbf{Contrastive Language-Image Pairing}
\\
A CLIP-like joint-embedding model usually consists of two main components: an image encoder and a text encoder and is trained with $N$ paired instances $\{(x_i, t_i)\}_{i=1}^N$, where $x_i$ is the image and $t_i$ represents the associated text of the pair with index $i$. The image encoder, denoted as $f_x(x_i)$, transforms an input image $x_i$ into an intermediate image embedding. This embedding is then projected into a shared embedding space in $\mathbb{R}^d$ with the image projector $g_x$, resulting in $u_i = g_x(f_x(x_i))$. 
Similarly, for the textual counterpart, the text encoder $f_t(t_i)$ processes an input text $t_i$ to produce an intermediate text embedding $y_i$, which is then mapped to the same shared space in $\mathbb{R}^d$ by the text projector $g_t$, giving $v_i = g_t(f_t(t_i))$.
The collections of embedding vectors are denoted by $U=[u_1, u_2, \cdots ,u_N]$ and $V=[v_1, v_2, \cdots ,v_N]$.

\begin{figure}[h!] % The figure environment
    \centering % Center the image
    \includegraphics[width=\linewidth]{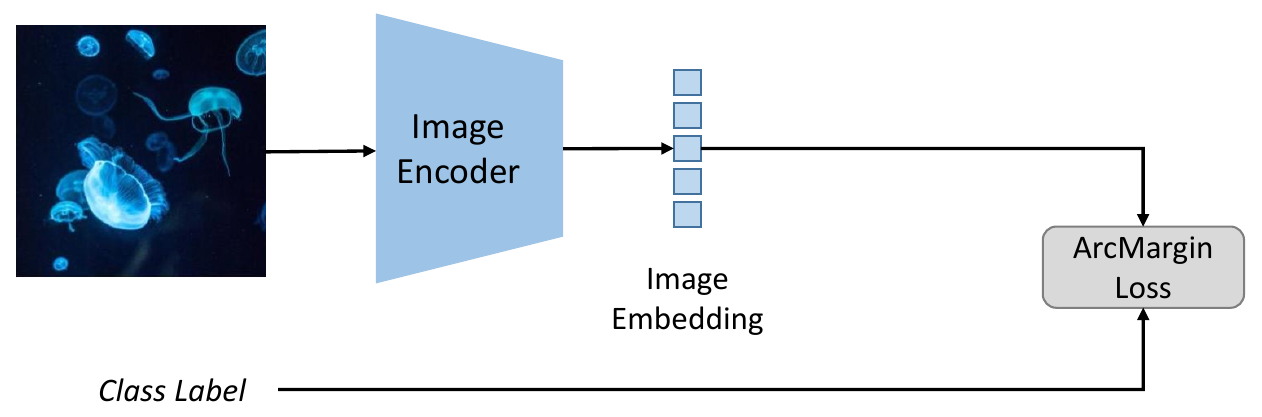} % Include the image
    \caption{General-Purpose-Retrieval Fine-Tuning: The image projector is discarded and the image encoder is solely fine-tuned with the ArcMargin loss on a dataset consisting of 22.6 million images from 168 thousand classes, with one class per image.} % Add a caption to the image
    \label{fig:gprft} % Label for referencing
\end{figure}

To optimize this joint embedding space, the model employs a contrastive loss function known as InfoNCE (Noise-Contrastive Estimation) \cite{InfoNCE}. This loss function aims to minimize the distance between the correct pairs of text and image embeddings in the shared space while maximizing the distance between mismatched pairs. It can be represented as
% \begin{align}
% & \mathcal{L}_{\text{InfoNCE}} = \nonumber \\
% & -\frac{1}{N} \sum_{i=1}^N \log \frac{\exp(u_i^\top v_i / \tau)}{\sum_{j=1}^N \exp(u_i^\top v_j /\tau)} 
%  - \frac{1}{N} \sum_{i=1}^N \log \frac{\exp(u_i^\top v_i / \tau)}{\sum_{j=1}^N \exp(u_j^\top v_i / \tau )}
% \label{eq:infosce}
% \end{align}
\begin{equation}
\mathcal{L}_{\text{InfoNCE}} = 
-\frac{1}{N} \sum_{i=1}^N \log \frac{\exp(u_i^\top v_i / \tau)}{\sum_{j=1}^N \exp(u_i^\top v_j /\tau)} 
 - \frac{1}{N} \sum_{i=1}^N \log \frac{\exp(u_i^\top v_i / \tau)}{\sum_{j=1}^N \exp(u_j^\top v_i / \tau )}
\label{eq:infosce}
\end{equation}

where $\tau$ is a temperature parameter that scales the dot products of the exponentials and since the elements of $U$ and $V$ are $L^2$-normalized the dot products represent the cosine similarities of the embeddings.
\\
\\
\textbf{General-Purpose Retrieval Fine-Tuning}
\\
General-Purpose Retrieval (GPR) \cite{GPR-NNS} fine-tunes the image encoder $f_x(x_i)$ (the projector is discarded) with a combined single-label dataset consisting of ImageNet20k (without the ImageNet1k classes) \cite{ImageNet}, Google-Landmarks V2 \cite{GLv2}, AliProducts \cite{Ali}, iNat21 \cite{iNat} and VGGFaces2 \cite{VGGFace}, totaling in 22.6 million images from 168k classes (Fig. \ref{fig:gprft}). The authors introduced a benchmark specifically for general-purpose retrieval and k-NN classification formed from various domains and showed that fine-tuning a CLIP image encoder achieves the best retrieval results across multiple domains without the need of a specific domain adoption. 

The average benchmark performance of the CLIP encoder can be increased by about 10 percent points when fine-tuning with their combined dataset and using the ArcMargin loss \cite{ArcFace,GPR-NNS}. The mix of data from a wide range of domains with different class-distributions and class boundaries, e.g narrow boundaries in the VGGFaces2 data and wide class boundaries in the ImageNet20k data parts, forces the previously semantically trained image-encoder to focus on visual features to distinguish the 168k classes of the training dataset while maintaining the generalization achieved with CLIP pre-training. 

The ArcMargin loss is a modification used in deep learning to enhance the discriminative power of features extracted by neural networks, especially in narrow class-boundary tasks like face recognition. This approach involves adjusting the angle between the feature of each sample and the corresponding weights of the samples class-vector in the classification layer. The main objective is to introduce a margin between classes within the angular space, thereby improving the separation of classes. The ArcMargin loss can be expressed as follows:
\begin{equation}
\mathcal{L}_{\text{ArcMargin}} = -\frac{1}{N} \sum_{i=1}^N \log \frac{\exp(s \cdot \cos(\theta_{i, y_i} + m))}{\exp(s \cdot \cos(\theta_{i, y_i} + m)) + \sum_{j \neq y_i} \exp(s \cdot \cos(\theta_{i, j}))}
\label{eq:arcmargin}
\end{equation}
where $N$ is the number of samples in the batch, $y_i$ denotes the true-class index for the $i^{\text{th}}$ sample,
and $\theta_{i, k}$ refers to the angle between the feature of the $i^{\text{th}}$ sample and the weight vector of the class index $k$.
%
%$\theta_{y_i}$ is the angle between the feature of the $i$-th sample and the weight corresponding to its true class at $y_i$, and $\theta_j$ refers to the angle between the feature of the $i$-th sample and the weight vector of other classes at index $j$. 
%
The parameter $s$ represents a scaling factor applied to the cosine of these angles, and $m$ is the margin added to the angle of the true class. 
$\cos(\theta_{y_i} + m)$ effectively pushes the decision boundary away from the feature vectors of the true class, enhancing model robustness. 

\subsection{Two-Stage Fine-Tuning}

The first proposed method includes two separate fine-tuning stages. The first stage optimizes the image-encoder in the same way as in the previously described GPR fine-tuning approach using the same combination of training data and the ArcMargin loss function. The fine-tuned image-encoder produces embeddings that capture more visual information of the images and performs better across a wide range of image-based similarity search tasks. Since \cite{GPR-NNS} has shown that image-to-image retrieval performance benefits from discarding $g_x$, we do the same for a fair comparison.

\begin{figure}[h] % The figure environment
    \centering % Center the image
    \includegraphics[width=\linewidth]{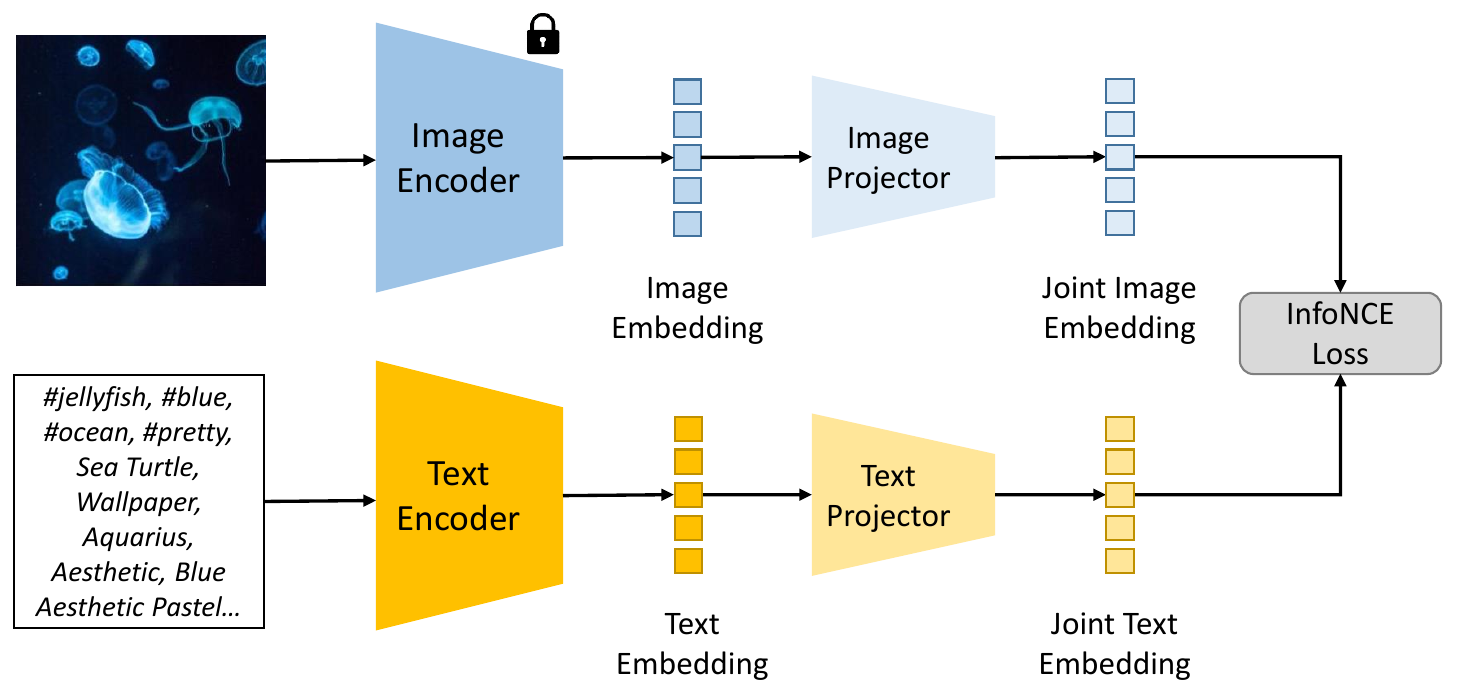} % Include the image
    \caption{Re-Alignment Fine-Tuning: The fine-tuned image-encoder is locked and the other three parts of the model are trained to re-align the text embeddings with the image embeddings. An image-captions paired dataset is used in this step.} % Add a caption to the image
    \label{fig:re-a} % Label for referencing
\end{figure}
Next, a re-alignment fine-tuning of the image and text embeddings is introduced with a image-text pair dataset. The procedure is shown in Fig. \ref{fig:re-a}. Since the first stage only fine-tuned the image-encoder $f_x$, in the second step the image-projector $g_x$, the text-encoder $f_t$, and the text-projector $g_t$ have to be re-trained to achieve an optimal alignment in the joint embedding space of $u \in \mathbb{R}^d$ and $v \in \mathbb{R}^d$. Similar to the work presented in LiT \cite{lit}, $f_x$ is locked during this training phase, i.e. the weights are not updated during back-propagation. 
This allows to extract all image embeddings prior to the re-alignment fine-tuning, which significantly reduces the memory load during training and enables the usage of large batch-sizes, which is beneficial for contrastive text-image training \cite{CLIP,SigLIP}. 
In this stage, the InfoNCE loss (Eq. \ref{eq:infosce}) is used. This method is denoted as 2SFT (two-stage fine-tuning) in the upcoming experiment section. The first stage will be called GPR-FT (general-purpose retrieval fine-tuning) and the second stage Re-A (re-alignment).

\subsection{General-Purpose Retrieval Fine-Tuning with Multi-Caption Image Pairings }
\begin{figure}[ht] % The figure environment
    \centering % Center the image
    \includegraphics[width=1.0\linewidth]{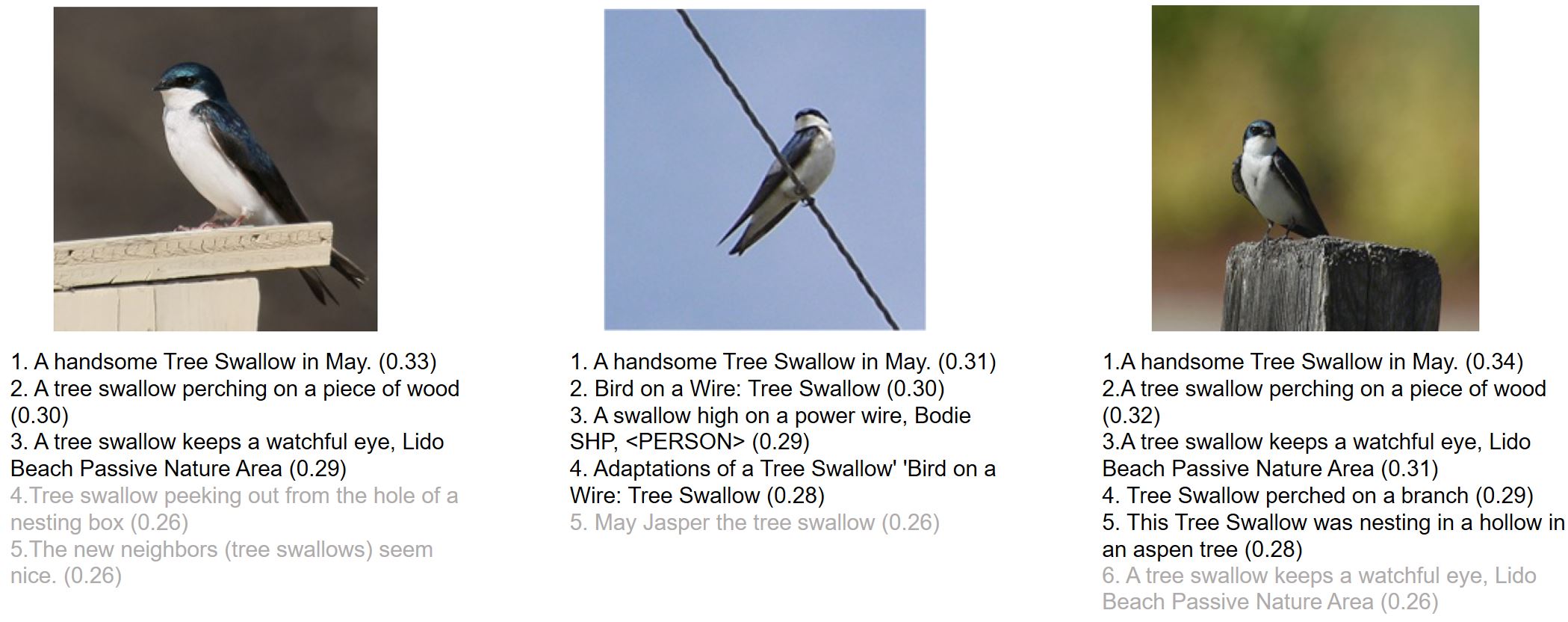} % Include the image
    \caption{Generated pseudo-captions for three images from one example class of the GPR-FT training set. Even though the texts are redundant, the combination of several captions succeeds in describing the content of the images.} % Add a caption to the image
    \label{fig:captions} % Label for referencing
\end{figure}

In contrast to the first method, we now try to incorporate the text-image pairing approach directly to the general-purpose retrieval fine-tuning with the goal to create an joint-embedding space with optimal image-retrieval while maintaining the alignment of the image and text embeddings. However, since the fine-tuning with the ArcMargin loss is performed using a single-class per image dataset without any available text-captions, the CLIP loss cannot be used in this fine-tuning strategy directly. 
\\
\\
\textbf{Pseudo-Captioning}
\\
We propose to employ a pseudo-captioning pipeline that utilizes a large collection of image-text pairs, specifically the Conceptional 12M dataset (CC12M) \cite{CC12M}. 
First, we extract image-embeddings for all images of the GPR training dataset with a pre-trained CLIP model and second, we extract text-embeddings for all captions available in CC12M. Next, we perform a similarity search with each of the image-embeddings to retrieve the k-nearest-neighbors from the text-embedding collection with $k=10$. Similar to previous work \cite{MOFI}, we subsequently discard all retrieved captions with a cosine similarity smaller than 0.27. This leads to a single-class-multi-caption-image paired collection, consisting of $N$ elements $\left\{(x_i, y_i, \{t_{i, 1}, t_{i, 2}, \ldots, t_{i, T_i}\})\right\}_{i=1}^N$, with $x_i$ being the image, $y_i$ being the true-class label and $\{t_{i, 1}, t_{i, 2}, \ldots, t_{i, T_i}\}$ being the collection of $T_i \leq k$ pseudo-captions with similarity score greater than 0.27. Figure \ref{fig:captions} shows the generated captions for three images from one example class of the training dataset. It is apparent, that even though the pseudo-captions are noisy and repetitive, the combination of multiple captions manages to describe the images content and has low intra-class variance. 

\noindent\textbf{ArcMargin Loss for Multi-Caption Image Pairs}
\\
To make use of the generated single-label-multi-caption-image tuples during the general-purpose image retrieval-finetuning, we propose an extension of the ArcMargin loss that utilizes the pseudo-captions to maintain joint-embedding alignment during the image-encoder optimization. 

\begin{figure}[t!] % The figure environment
    \centering % Center the image
    \includegraphics[width=\linewidth]{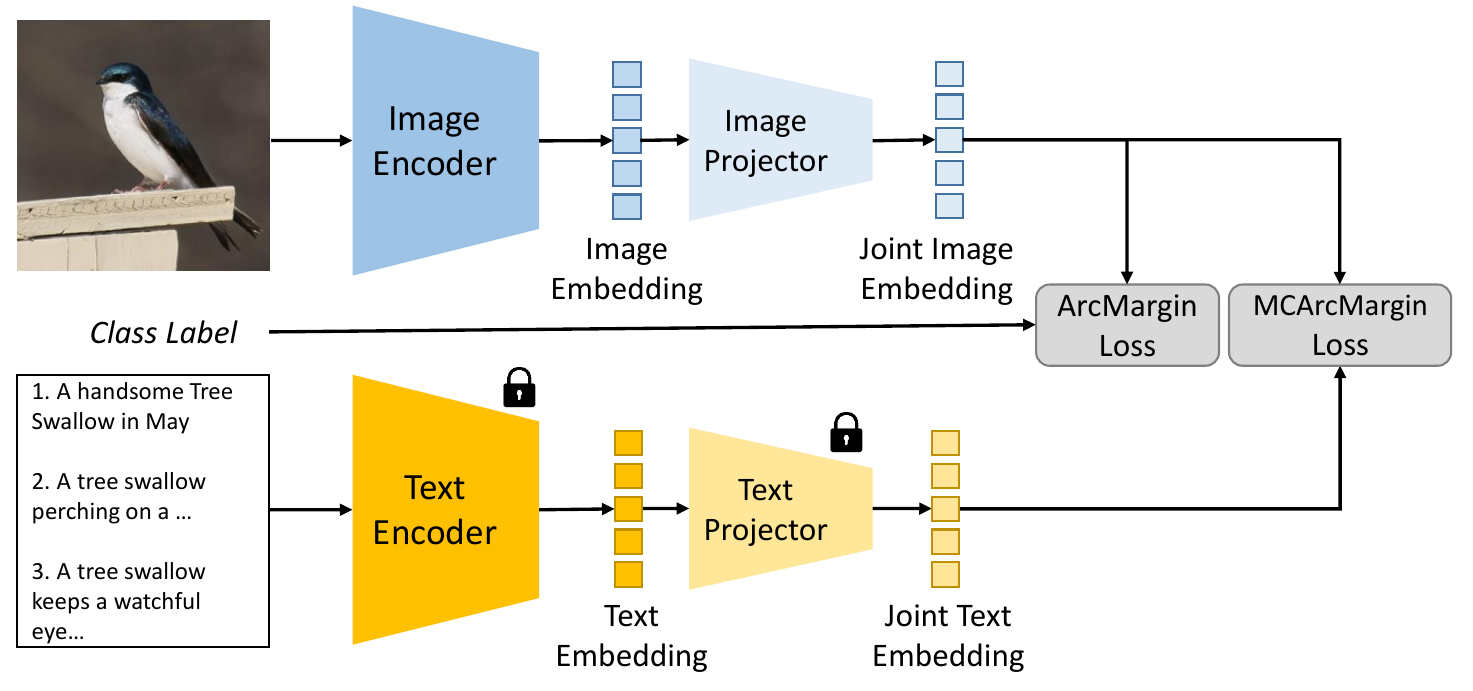} % Include the image
    \caption{Multi-Caption Image Pairing Fine-Tuning: This method utilizes the generated pseudo-captions to the train the image encoder and projector while keeping the text encoder and projector locked to maintain alignment of the joint-embedding space in a single fine-tuning run. In addition to the single-label training with the ArcMargin loss, the introduced Multi-Caption-ArcMargin loss is used in this step.} % Add a caption to the image
    \label{fig:mcip} % Label for referencing
\end{figure}
The ArcMargin loss function as given in Eq. \ref{eq:arcmargin} can be extended to the Multi-Label-ArcMargin (MLArcMargin) loss to enable multi-label training as follows:
\begin{align}
& \mathcal{L}_{\text{MLArcMargin}} = \notag \\
& -\frac{1}{N} \sum_{i=1}^N \frac{1}{L_i} \sum_{l=1}^{L_i} \log \frac{\exp(s \cdot \cos(\theta_{i, y_{i_l}} + m))}{\exp(s \cdot \cos(\theta_{y_{i_l}} + m)) + \sum_{j \neq y_{i_l}} \exp(s \cdot \cos(\theta_{i,j}))}
\label{eq:mcarcmargin}
\end{align}
where $N$ is the number of images in the batch, $L_i$ is the number of labels of the $i^{\text{th}}$ sample and $\theta_{i, y_{i_l}}$ is the angle between $u_i$ (the image-embedding of sample at index $i$) and the weight vector representing the $l^{\text{th}}$ class of the sample at index $i$.

However, since in our case we do not have multiple labels but captions per image, we modify the Multi-Label-ArcMargin loss to the Multi-Caption-ArcMargin (MCArcMargin) loss as given by
%
% \begin{align}
% \mathcal{L}_{\text{MCArcMargin}} = & -\frac{1}{N} \sum_{i=1}^N \frac{1}{T_i} \sum_{l=1}^{T_i} \text{logExp} \notag \\
% \text{where } \text{logExp} = & \log \left(\frac{\exp(s \cdot \cos(\theta_{t_{i_l}} + m))}{\exp(s \cdot \cos(\theta_{t_{i_l}} + m)) + \sum_{j \neq t_{i_l}} \exp(s \cdot \cos(\theta_j))}\right)
% \label{eq:mtargmargin}
% \end{align}
%
% Versuch von Klaus:
\begin{align}
& \mathcal{L}_{\text{MCArcMargin}} = \notag \\
  & -\frac{1}{N} \sum_{i=1}^N \frac{1}{C_i} \sum_{c=1}^{C_i} \log \frac{\exp(s \cdot \cos(\theta_{i, t_{i, c}} + m))}{\exp(s \cdot \cos(\theta_{i, t_{i, c}} + m)) + \sum_{t \neq t_{i, c}} \exp(s \cdot \cos(\theta_{i, t}))}
\label{eq:mtargmargin}
\end{align}
where $C_i$ is the number of captions generated for the image at index $i$, $\theta_{i, t_{i, c}}$ is the angle between the image-embedding $u_i$ and the text-embedding $v_{i, c}$ (the $c^{\text{th}}$ generated caption of sample $i$) and $\theta_{i, t}$ refers to the angle between the feature of the $i^{\text{th}}$ sample and the text-embedding of caption $t$ in the batch other than $t_{i, c}$. $m$ and $s$ are the same hyper-parameters as in the original ArcMargin loss function (Eq. \ref{eq:arcmargin}). This loss function ensures that the angles between an image-embedding and the text-embeddings of the generated captions for this image are small and enforces a larger angle for the text-embeddings of other images in the batch. 

Additionally, we also compute the the standard ArcMargin loss for the class-labels, akin to the previous approach, setting the final loss objective to:

\begin{equation}
\mathcal{L} = \lambda_1 \mathcal{L}_{\text{ArcMargin}} + \lambda_2 \mathcal{L}_{\text{MCArcMargin}}
\label{eq:finalLoss}
\end{equation}

where $\lambda_1$ and $\lambda_2$ are two hyper-parameters that allow to modify the weight of the respective loss function.

In order to maintain the joint-embedding alignment, the image-encoder and the image-projector are simultaneously fine-tuned, while the text-encoder and text-projector remain locked. This method is further denoted as MCIP (Multi-Caption-Image-Pairing).

\section{Experiments}
This section presents the experiments to evaluate our two proposed methods. We used the AdamW optimizer \cite{AdamW} in all fine-tuning runs and report results for the best learning rate, which varies for each of the evaluated models. Additionally, we apply weight decay with a factor of $10^{-3}$. As for the MCIP approach, we use $\lambda_1=0.5$ and $\lambda_2=0.5$. All experiments where performed on two NVIDIA A100 GPUs.

\subsection{Evaluation}
Since the goal is to optimize CLIP-like models to extract robust embeddings and to perform well in domain-independent similarity search tasks like content-based image retrieval, k-NN classification, zero-shot classification and text-to-image retrieval, we introduce a benchmark that consists of several evaluation sets for each of the tasks. 
\\
\\
\textbf{Content-Based Image Retrieval}
\\
Starting with content-based image retrieval, we evaluate the image-embeddings with GPR1200 \cite{GPR1200}, a challenging evaluation set, consisting of 1200 categories from various domains. It has been shown that the GPR1200 mean-average-precision (mAP) has a very high correlation to the average of nine different evaluation sets \cite{GPR-NNS}. Additionally, we add the INSTRE dataset \cite{INSTRE}, an evaluation set with mixed domains, the CUB200 dataset \cite{CUB_200_2011} and the ROxford dataset \cite{RParisOxford} on the medium setting. We apply the respective evaluation protocol for each dataset and report mAP values for INSTRE, GPR1200 and ROxford and Recall@1 values for CUB200.
\\
\\
\\
\textbf{k-NN Classification}
\\
The main advantage of k-NN classification over linear-probing is that the model has not have to be fine-tuned to handle the specific class-distribution of an evaluation set and can directly be used by extracting the image-embeddings of the training split from a dataset. For each test-image, the $k$ nearest neighbors are then simply retrieved from previously extracted training-data embeddings and the predicted class is the one with the most occurrences in the retrieved neighbors collection. We set $k=21$ in this evaluation. The included testsets are ImageNet1k-Validation (IN-V) \cite{ImageNet}, one of the most prominent evaluation sets in computer-vision, the ImageNet-Sketch dataset (IN-S) \cite{ImageNetSketch}, a variant of ImageNet1k, with the same class-distribution, but domain-shifted images, namely sketches, the ImageNet-Adversarial dataset (IN-A) \cite{ImageNet-A}, similar to ImageNet1k, but with images, where the desired classes objects are often showed together with other items. Last, we add the FGVCAircraft (FGVCA) dataset \cite{Aircraft}, containing images from 100 different airplanes and helicopters. While the three ImageNet variants all use the ImageNet1k training-set to perform k-NN classification, the FGVCAircraft dataset brings its own training-split. All evaluations report the classification accuracy. 
\\
\\
\textbf{Zero-Shot Classification}
\\
Zero-Shot classification is particularly useful in scenarios where not enough labeled images are available for performing k-NN classification and represents one of the most common applications of CLIP-like joint-embedding models. All available classes of an evaluation set are parsed with a prompt similar to "an image of <class-name>" and are embedded in the joint-embedding space with the text-encoder. Next, each image is embedded with the image-encoder and the cosine similarity to each of the class-text-embeddings is computed. The predicted class is the one with the highest similarity to the encoded image-embedding. 
We evaluate the zero-shot classification accuracy for the same four evaluation sets as introduced in the k-NN evaluation with the addition of ObjectNet (ON) \cite{ObjectNet}, a large-scale dataset where object backgrounds, rotations, and views are random. 
\\
\\
\textbf{Text-to-Image Retrieval}
\\
Text-to-image retrieval is the task of finding an image according to a given textual description. Text-to-image retrieval evaluation sets therefore consist of text-image pairs and CLIP-like models use the text-embeddings to search in a collection of previously extracted image-embeddings. 
The metrics are either Recall@1, i.e. is the respective image the one with highest similarity when searching with the respective text-vector - and Recall@5, i.e. is the respective image among the top-5 retrieved results. 
We evaluate text-to-image retrieval performance on Flickr30k (F30k) \cite{Flickr30k} and the MS-COCO-Caption evaluation sets (COCO) \cite{MSCoco} and report the Recall@5 metric, since it is closer to real-world applications compared to Recall@1.

\subsection{Results}

\begin{table*}[ht]
    \resizebox{1.0\textwidth}{!}{%
    \centering
    \setlength{\tabcolsep}{1pt}
    \small
    % \begin{tabular} {c|ccc|cccc|cccc|cccc|cc|c}
    \begin{tabular} {p{1.2cm}|p{0.28cm}p{0.28cm}p{0.28cm}|cccc|cccc|cccc|cc|c}
        \multicolumn{4}{c|}{} & \multicolumn{4}{c|}{\textbf{I2I}} & \multicolumn{4}{c|}{\textbf{k-NN}} & \multicolumn{4}{c|}{\textbf{Zero-Shot}} & \multicolumn{2}{c|}{\textbf{T2I}} & \multicolumn{1}{c}{} \\
        
        \rotatebox{0}{\textbf{Model}} & \rotatebox{90}{\textbf{GPR-FT}}  & \rotatebox{90}{\textbf{MCIP}} & \rotatebox{90}{\textbf{Re-A}} & \rotatebox{90}{\textbf{GPR1200}} & \rotatebox{90}{\textbf{INSTRE}} & \rotatebox{90}{\textbf{CUB200}} & \rotatebox{90}{\textbf{ROxford}} &  \rotatebox{90}{\textbf{IN-V}} & \rotatebox{90}{\textbf{IN-S}} & \rotatebox{90}{\textbf{IN-A}} &
    \rotatebox{90}{\textbf{FGVCA}} & \rotatebox{90}{\textbf{IN-V}} & \rotatebox{90}{\textbf{IN-S}} & \rotatebox{90}{\textbf{IN-A}} & \rotatebox{90}{\textbf{ON}} &
    \rotatebox{90}{\textbf{Flickr30k}} & \rotatebox{90}{\textbf{MS-COCO}} & \rotatebox{90}{\textbf{AVG}}
        \\ 
        \hline
       
        OpenAI &  &  &  & 76.0 & 77.5 & 77.0 & 43.4 & 78.2 & 52.7 & 46.4 & 51.3 & 76.6 & 61.0 & 77.6 & 72.0 & 89.0 & 61.7 & 67.2 \\
        OpenAI & \cmark &   &   & \textbf{87.1} & \textbf{85.0} & \textbf{88.2} & 67.3 & \textbf{84.2} & \textbf{60.5} & \textbf{61.2} & \textbf{57.8} & 74.5 & 60.8 & 76.2 & 74.5 & 84.9 & 57.2 & 68.4 \\
        OpenAI & \cmark &   & \cmark & 86.7 & 84.4 & 87.9 & 66.4 & 83.8 & 60.4 & 60.9 & 57.1 & 77.2 & 61.5 & 77.9 & 74.6 & 89.8 & 64.6 & 73.8 \\
        OpenAI &   & \cmark &   & 87.0 & 84.8 & 88.0 & \textbf{67.4} & 84.1 & 60.4 & 61.0 & 57.7 & 77.0 & 61.1 & 78.3 & 74.5 & 90.1 & 65.6 & 74.1 \\
        OpenAI &   & \cmark & \cmark & 87.0 & 84.8 & 88.0 & \textbf{67.4} & 84.1 & 60.4 & 61.0 & 57.7 & \textbf{77.3} & \textbf{61.6} & \textbf{78.4} & \textbf{74.6} & \textbf{90.2} & \textbf{66.1} & \textbf{74.2}
        \\ 
        \hline
        LAION &   &   &   & 77.2 & 80.2 & 82.2 & 50.8 & 79.5 & 61.0 & 43.8 & 59.3 & 79.2 & 68.0 & 69.6 & 74.3 & 91.6 & \textbf{70.3} & 70.5 \\
        LAION &   & \cmark & \cmark & \textbf{87.5} & \textbf{82.1} & \textbf{90.2} & \textbf{72.3} & \textbf{83.8} & \textbf{65.8} & \textbf{61.5} & \textbf{63.1} & \textbf{80.2} & \textbf{68.2} & \textbf{75.2} & \textbf{79.7}  & \textbf{91.7} & 69.8 & \textbf{76.5} \\
        \hline
        SigLIP &   &   &   & 78.2 & 84.1 & 82.4 & 51.3 & 85.2 & 67.1 & 59.7 & 71.8 & \textbf{83.1} & 74.5 & 82.5 & 76.9 & 93.5 & \textbf{76.7} & 76.2 \\
        SigLIP &   & \cmark & \cmark & \textbf{89.0} & \textbf{85.1} & \textbf{91.2} & \textbf{70.5} & \textbf{86.4} & \textbf{70.2} & \textbf{68.7} & \textbf{72.3} & 82.9 & \textbf{74.7} & \textbf{84.4} & \textbf{80.3} & \textbf{95.0} & 75.3 & \textbf{80.4}
        
    \end{tabular}
    } % resizebox
    \\[1 ex] % add extra space to caption
   % \caption{\small I2I: Image-to-Image retrieval (mAP/R@1), k-NN: k-NN classification (accuracy) for $k=21$, Zero-Shot: zero-shot classification (accuracy), T2I: Text-to-Image retrieval (R@5). First section: Results for each individual step of our proposed methods when applied to the original OpenAI CLIP model (ViT-L@336). Second section: Results for the Datacomb trained ViT-L by LAION. Last section: Results for the SigLIP ViT-SO400M model.  }

\caption{\small \textbf{I2I}: Image-to-Image retrieval (mAP/R@1),  \textbf{k-NN} classification (accuracy), $k=21$,  \textbf{Zero-Shot} classification (accuracy), \textbf{T2I}: Text-to-Image retrieval (R@5). 
    First section: Results for each individual step of our proposed methods when applied to the original OpenAI CLIP model (ViT-L@336).
    Second section: Results for the Datacomb trained ViT-L by LAION. 
    Last section: Results for the SigLIP ViT-SO400M model.
    }
    \label{tab:results1}
\end{table*}

We begin by examining the impact of each individual step in the two-stage fine-tuning approach (GPR-FT + Re-A), as well as the multi-caption-image-pairing (MCIP) fine-tuning, both with and without an additional re-alignment step. We compare these results to the baseline model, which is the original OpenAI ViT-L@336 \cite{CLIP}, a vision transformer \cite{VIT} that processes images with a resolution of 336x336 pixels. Table \ref{tab:results1} shows the results for each of the introduced evaluation sets. 

The model achieves the highest scores among the tasks of image-to-image retrieval and k-NN classification after the GPR fine-tuning, but, as expected, achieves the lowest evaluation scores across the text-based tasks in this state. Since only the image encoder was fine-tuned with a discarded image-projector, the image-embeddings are not aligned with the image-projector that has to be used to perform the text-based evaluations, which leads to significantly lower results. The re-alignment fine-tuning with the 12 million image-text pairs from the CC12M dataset leads to a strong improvement in all of the text-based tasks (Zero-Shot and T2I), however, the image-based evaluation scores are slightly reduced compared to the previous model state. 

Next, the model was trained with the introduced multi-caption-image-pairing approach (MCIP). It achieves higher scores across all of the evaluation tasks compared to the two-stage-fine-tuning method after a single training run. However, it can be observed that the performance is further slightly increased with an additional re-alignment fine-tuning step. This is performed as described in Fig. \ref{fig:re-a} with the small adjustment of additionally locking the image projector, since this part was also trained in the MCIP step. 
This results show that the MCIP+Re-A fine-tuning leads to better results compared to the naive two-stage-fine-tuning, however the increase in average performance is rather low (73.8 vs 74.2). MCIP requires the generation of pseudo-captions and therefore has a slightly higher computational overhead.

\begin{figure}[ht] % The figure environment
    \centering % Center the image
    \includegraphics[width=\linewidth]{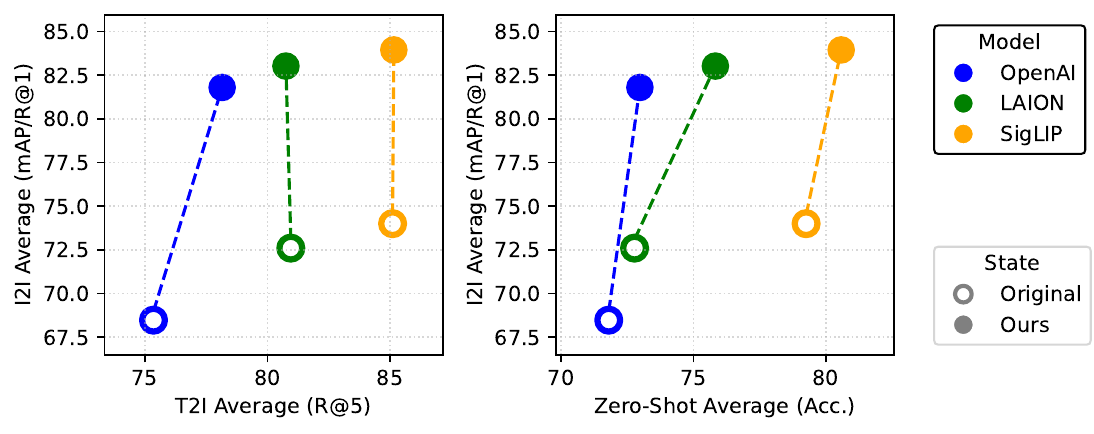} % Include the image
    \caption{Highlighted results for three different models fine-tuned with our proposed MCIP approach. In all three cases image-based similarity search could significantly be improved while maintaining text-to-image retrieval qualities and additionally improving zero-shot classification capabilities. }
    \label{fig:results} % Label for referencing
\end{figure}

Furthermore, we evaluated our methods with two additional models. First, a Vit-L variant that was trained with the Datacomp dataset \cite{datacomp}, even though smaller in size, it achieves better evaluation results compared to a ViT-H, trained with LAION5B \cite{Laion5B}. Second, the ViT-SO400M, a high-performance vision transformer where the number of parameters was optimized to 400 million \cite{SigLIP}. This model was trained with the webli dataset \cite{SigLIP} and with the Sigmoid-Language-Image-Pairing loss function (SigLIP) and reaches the highest scores in our evaluation in the baseline state. Since our MCIP method overall performed better than our 2SFT, we only report results for the multi-caption-image-pairing fine-tuning. 
While our proposed method leads to a significant increase in both image-based similarity search tasks, text-based zero-shot classification was only improved with the ViT-L model for all datasets and only for some with the ViT-SO400M model. Interestingly, the text-to-image retrieval quality was increased on the Flickr30k dataset and slightly decreased on the MS-COCO evaluation set after fine-tuning with our MCIP approach. Nevertheless, this experiments show that our method is able to optimize several CLIP-like models for image-retrieval while also maintaining or even improve the performance in text-based tasks like zero-shot classification and text-to-image retrieval. Fig. \ref{fig:results} shows some highlighted results in a 2D scatter plot.

\section{Conclusions}

In this paper, we addressed the challenge of optimizing CLIP models for image retrieval while preserving the joint-embedding alignment crucial for effective text-based search tasks. Through a detailed exploration of two novel methodologies, we demonstrated how fine-tuning CLIP models can enhance image-based similarity search capabilities without compromising text-to-image retrieval and zero-shot classification performance.

Our first proposed method, the two-stage fine-tuning (2SFT) approach, involves initially fine-tuning the image encoder for improved retrieval, followed by a re-alignment of the text encoder. This method, while effective, showed a slight reduction in text-based tasks performance after re-alignment. The second method, which integrates multi-caption-image pairing (MCIP) during the retrieval optimization phase, resulted in better overall performance. The MCIP approach, particularly when followed by a re-alignment step, achieved higher scores across all evaluated tasks, proving to be a more comprehensive solution.

We validated our methods on various models, including the original \hbox{OpenAI} CLIP ViT-L@336, a Datacomp trained ViT-L model, and the SigLIP ViT-SO400M model. Our experiments revealed significant improvements in image retrieval benchmarks like GPR1200 and INSTRE, while maintaining or even enhancing the performance in zero-shot classification and text-to-image retrieval tasks.

The results confirm that our proposed methodologies not only optimize CLIP models for diverse image retrieval scenarios but also maintain the robustness required for text-based search tasks. This dual optimization is critical for real-world applications where both modalities are often required to work seamlessly together.

Future work will focus on expanding the pseudo-caption generation process to include more diverse and contextually rich text descriptions which could provide even better alignment and retrieval performance. Additionally, we will explore the idea of using automated machine translation of the pseudo-captions to boost specific language performance in text-to-image retrieval.

The code and optimized model weights for our experiments are available at: \repourl.

\bibliographystyle{splncs04}
\bibliography{references}
\end{document}